\documentclass[conference]{IEEEtran}

\usepackage[utf8]{inputenc}
\usepackage{amsmath, amssymb, amsthm}
\usepackage{graphicx}
\usepackage{hyperref}
\usepackage{cite}
\usepackage{float}
\usepackage{algorithm}
\usepackage{algorithmic}
\usepackage{footnote}
\usepackage{booktabs}
\usepackage{tabularx}
\usepackage{afterpage}

\title{FDQN: A Flexible Deep Q-Network Framework for Game Automation}
\author{\IEEEauthorblockN{Prabhath Reddy Gujavarthy}
\IEEEauthorblockA{
    Department of Computer Engineering\\
    San José State University (SJSU)\\
    San Jose, CA, USA\\
    \texttt{prabhathreddy.gujavarthy@sjsu.edu}}
}

\begin{document}

\maketitle

\begin{abstract}
In reinforcement learning, it is often difficult to automate high-dimensional, rapid decision-making in dynamic environments, especially when domains require real-time online interaction and adaptive strategies such as web-based games. This work proposes a state-of-the-art Flexible Deep Q-Network (FDQN) framework that can address this challenge with a self-adaptive approach that is processing high-dimensional sensory data in realtime using a CNN and dynamically adapting the model architecture to varying action spaces of different gaming environments and outperforming previous baseline models in various Atari games and the Chrome Dino game as baselines. Using the epsilon-greedy policy, it effectively balances the new learning and exploitation for improved performance, and it has been designed with a modular structure that it can be easily adapted to other HTML-based games without touching the core part of the framework. It is demonstrated that the FDQN framework can successfully solve a well-defined task in a laboratory condition, but more importantly it also discusses potential applications to more challenging real-world cases and serve as the starting point for future further exploration into automated game play and beyond.
\end{abstract}

\IEEEpeerreviewmaketitle

\afterpage{
\begin{figure*}[htbp]
    \centering
    \includegraphics[width=\textwidth, height=0.4 \textwidth]{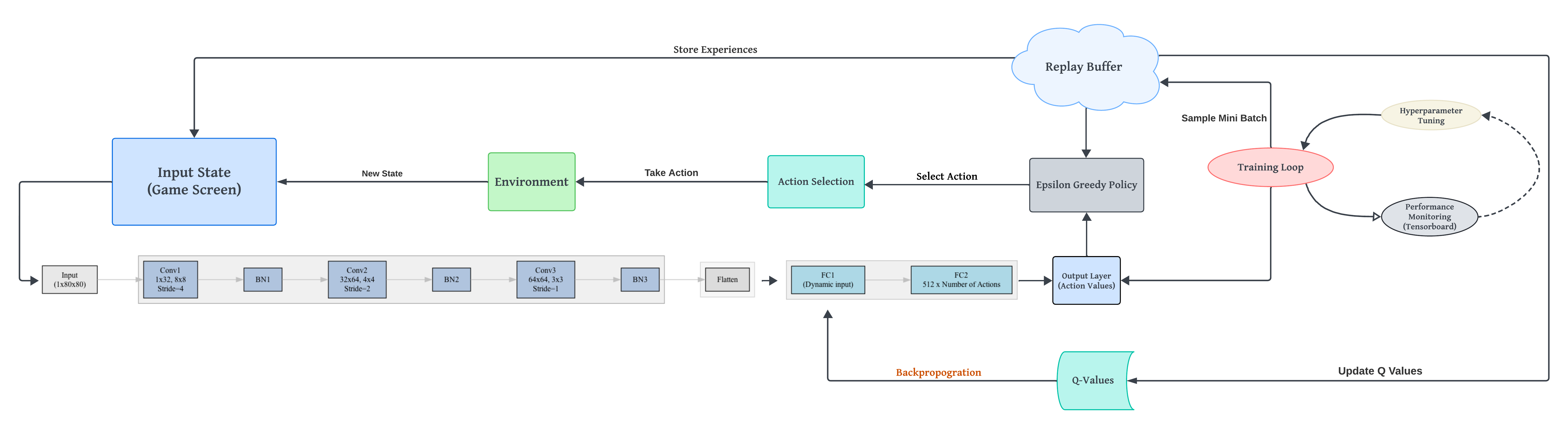} 
    \caption{Project Architecture}
    \label{fig:architecture}
\end{figure*}
}

\section{Introduction}
\label{sec:introduction}

Reinforcement learning is a powerful technique for training Machine learning models (widely called Agents) to make sequential decisions in a system, also known as environment. Among the popular architectures in training RL models, the Deep Q-Networks (DQNs) are very suitable at enabling these agents to learn from high-dimensional sensor data, like images and video. This paper presents a Flexible DQN(FDQN) framework, designed to enable agents to navigate through dynamic environments. This architecture is designed to work on multiple Atari games and can also act on simple internet games like Chrome Dino.

Each Atari game has unique challenges and difficulty levels. There is a need for real-time decision making during the agent-environment interaction, which makes it a great place to test out this architecture. The framework is built keeping in mind the varying action space(A) of different environments. The visual input on the screen is processed through a CNN to capture the visual features, and an epsilon greedy policy is derived based on the interactions. A large replay buffer is placed to stabilize the training by mitigating the correlation between consecutive experiences, and ensures the agent navigates efficiently through the state space. 

This work provides a generalized RL framework that is easy to adapt and extend to any HTML based game, and offers an in-depth analysis on the architectural choices for optimal agent performance. Sections \ref{sec:methodology}, \ref{sec:experiments} will go through the architecture, experimental setup and the results obtained. Sections \ref{sec:discussion} and \ref{sec:conclusion} discuss the findings and conclude the paper.

\section{Related Work}
\label{sec:related_work}

Reinforcement learning has shown lots of promise at solving complex decision-making problems across robotics, autonomous systems and even in gaming industries. Deep Q-Networks (DQNs) have been particularly effective in learning control policies directly from high-dimensional sensory inputs, such as raw pixels from video games.

The work by Mnih et al. \cite{mnih2015human} introduced the DQN, and combines Q-learning with CNNs to play Atari games almost at human-level performance. The way the modern deep learning networks can handle high-dimensional input spaces such as video games provided a breakthrough in building these intelligent systems that are able to train themselves simply based on direct interaction. Double DQN \cite{van2016deep}, Dueling DQN \cite{wang2016dueling}, Prioritized Experience Replay \cite{schaul2015prioritized}, and other optimization techniques have greatly improved the stability and efficiency of DQNs.

OpenAI built a platform API called Gym to provide a standardized platform for training and benchmarking RL algorithms \cite{brockman2016openai}. RL has been widely applied to the Chrome Dino game as a benchmark like the work done by Luong \cite{luong2020chrome}. Singh \cite{singh2021deep} trained DQNs based on the environments, where the state and action spaces are known. These advancements have led to the development of various RL algorithms capable of handling both discrete and continuous action spaces.

This work builds on top of these ideas, developing a simple yet adaptable DQN framework which can be easily trained to adapt to multiple games. Unlike the previous studies that focused on specific games, this one offers a level of flexibility and extensibility, and can play multiple games using the same base version. I have also provided a set of well-tested hyperparameters and architectural choices for customizing it based on the user’s requirements and the complexity of the environment.

\section{Methodology}
\label{sec:methodology}

\subsection{Problem Definition}
\label{subsec:problem_definition}
The goal of this research is to develop a customizable Deep Q-Network (DQN) framework capable of automating web-based games with Chrome Dino as a baseline. The architecture allows the agent to select actions dynamically based on the environment. The agent interacts with this environment by observing states, selecting actions, and receiving rewards, aiming to learn an optimal policy $\pi^*$ that maximizes current and future rewards.

This problem is mathematically modeled as a Markov Decision Process (MDP) defined by the tuple $(S, A, P, R, \gamma)$ where:
\begin{itemize}
    \item $S$ is the state space,
    \item $A$ is the action space,
    \item $P(s'|s,a)$ is the state transition probability,
    \item $R(s,a)$ is the reward function, and
    \item $\gamma \in [0,1]$ is the discount factor.
\end{itemize}
The goal is to find a policy $\pi: S \to A$ to maximize the expected reward $\mathbb{E}[R_t | s_t, a_t]$.

\subsection{Deep Q-Network (DQN) Architecture}
\label{subsec:dqn_architecture}
The DQN agent uses a CNN architecture for observing the game state and approximating the Q-function, $Q(s,a; \theta)$, where $\theta$ represents the network parameters. The network architecture has multiple convolutional layers followed by fully connected layers, and the output layer provides the Q-values for all the possible actions at every given state.

The loss function for updating the network parameters is defined as:
\[
L(\theta) = \mathbb{E}_{(s,a,r,s') \sim \mathcal{D}} \left[ \left( r + \gamma \max_{a'} Q(s', a'; \theta^{-}) - Q(s, a; \theta) \right)^2 \right]
\]
where $\theta^{-}$ are the parameters of a target network, $\mathcal{D}$ is the replay buffer, and $r$ is the reward received after taking action $a$ from state $s$.

\subsection{Replay Buffer}
\label{subsec:replay_buffer}
A replay buffer is used to store the agent's experiences $(s, a, r, s')$. During the training, mini-batches of experiences are sampled uniformly from the replay buffer to break the correlation between consecutive experiences.

\subsection{Epsilon-Greedy Policy}
\label{subsec:epsilon_greedy_policy}
The agent balances exploration and exploitation using an epsilon-greedy policy. The exploration rate $\epsilon$ decays over time according to:
\[
\epsilon = \max(\epsilon_{\text{min}}, \epsilon_{\text{max}} \cdot \text{decay}^t)
\]
where $\epsilon_{\text{min}}$ and $\epsilon_{\text{max}}$ are the minimum and maximum exploration rates, respectively, and $t$ is the timestep.

\subsection{Double DQN}
\label{subsec:double_dqn}
To mitigate overestimation bias in Q-learning, Double DQN is utilized, which separates the action selection and evaluation:
\[
y = r + \gamma Q(s', \arg\max_{a'} Q(s', a'; \theta); \theta^{-})
\]
where $\theta$ and $\theta^{-}$ are the parameters of the online and target networks, respectively.

\subsection{Algorithm}
\label{subsec:algorithm}
The training procedure for the DQN agent is outlined in Algorithm \ref{alg:dqn}.

\begin{algorithm}
\caption{Training the DQN Agent}
\label{alg:dqn}
\begin{algorithmic}[1]
\STATE Initialize replay buffer $\mathcal{D}$
\STATE Initialize online network $Q(s, a; \theta)$ with random weights
\STATE Initialize target network $Q(s, a; \theta^{-})$ with weights $\theta^{-} \leftarrow \theta$
\FOR{each episode}
    \STATE Initialize state $s$
    \FOR{each step in episode}
        \STATE With probability $\epsilon$ select a random action $a$, otherwise select $a = \arg\max_a Q(s, a; \theta)$
        \STATE Execute action $a$ and observe reward $r$ and next state $s'$
        \STATE Store transition $(s, a, r, s')$ in $\mathcal{D}$
        \STATE Sample random mini-batch of transitions $(s_j, a_j, r_j, s_j')$ from $\mathcal{D}$
        \STATE Compute the target $y_j = r_j + \gamma \max_{a'} Q(s_j', a'; \theta^{-})$
        \STATE Perform gradient descent step on $(y_j - Q(s_j, a_j; \theta))^2$ with respect to $\theta$
        \STATE Update state $s \leftarrow s'$
    \ENDFOR
    \STATE Update the target network weights $\theta^{-} \leftarrow \theta$
\ENDFOR
\STATE Return the trained network $Q(s, a; \theta)$
\end{algorithmic}
\end{algorithm}

\subsection{Optimization Steps}
\label{subsec:optimization_steps}
The optimization of the DQN agent involves the following steps:
\begin{enumerate}
    \item \textbf{Initialization of Parameters}: The Online(Learner) and Target(Best Policy) networks were initialized with random weights to promote initial diversity in learning trajectories and mitigate any convergence to suboptimal policies.

    \item \textbf{Experience Replay Mechanism}: Implemented a Standard Replay Buffer that can store up to 1 million experiences.

    \item \textbf{Target Network Update}: Target network's parameters are updated periodically(not per every step), to stabilize the training updates while calculating temporal-difference error using consistent learning targets .

    \item \textbf{Gradient Descent Optimization}: Adam Optimizer was used to perform gradient descent and updating the online network's weights based on the computed loss between the predicted and target Q-values.
\end{enumerate}

\subsection{Customizable Framework}
\label{subsec:customizable_framework}
This framework is designed to be highly customizable, and adapt DQN agent to different environments with little-to-zero modifications. The key components that can be customized through this framework include:
\begin{itemize}
    \item Environment wrappers for different games.
    \item Network architecture for different state representations.
    \item Hyperparameter settings for different training policies.
\end{itemize}

\section{Experiments}
\label{sec:experiments}

This section presents the experimental setup, methodology, and results for evaluating the performance of the advanced DQN framework on various environments, including multiple OpenAI Gym environments and web-based games like Chrome Dino. The experiments were conducted to demonstrate the framework's adaptability and effectiveness across diverse tasks.

\subsection{Experimental Setup}
\label{sec:setup}

\subsubsection{Environments}
\label{subsec:environments}
The following environments were utilized for evaluating the DQN framework:

\subsubsection{Game Environments Descriptions}
\label{subsec:game_environments_descriptions}

\begin{itemize}
    \item \textbf{Chrome Dino}: A simple web-based game requiring the agent to jump over obstacles.
    \item \textbf{Breakout}: A classic arcade game where the agent controls a paddle to bounce a ball and break bricks.
    \item \textbf{Pong}: A two-player game where the agent controls a paddle to hit a ball and score against an opponent.
    \item \textbf{CartPole}: An environment where the agent balances a pole on a moving cart.
    \item \textbf{MountainCar}: An environment where the agent drives a car up a steep hill.
    \item \textbf{Assault}: An arcade shooter game where the agent aims to survive and score points by shooting enemies.
    \item \textbf{Frostbite}: A game where the agent builds an igloo while avoiding obstacles and enemies.
    \item \textbf{Pacman}: A maze arcade game where the agent navigates through the maze, eating pellets and avoiding ghosts.
    \item \textbf{Qbert}: A game where the agent hops around a pyramid of cubes to change their color while avoiding enemies.
    \item \textbf{Seaquest}: An underwater shooter game where the agent rescues divers and avoids enemies.
    \item \textbf{Space Invaders}: A classic arcade game where the agent shoots descending aliens.
\end{itemize}

\begin{table}[ht]
    \centering
    \caption{Action Sizes for Each Game}
    \begin{tabular}{cc}
        \toprule
        \toprule 
        \textbf{Game} & \textbf{Action Size} \\
        \midrule
        Chrome Dino & 2 \\
        Breakout & 4 \\
        Pong & 3 \\
        CartPole & 2 \\
        MountainCar & 3 \\
        Assault & 7 \\
        Frostbite & 4 \\
        Pacman & 4 \\
        Qbert & 4 \\
        Seaquest & 18 \\
        Space Invaders & 6 \\
        \bottomrule
    \end{tabular}
    \label{tab:actionsizes}
\end{table}

\subsubsection{Hyperparameter Tuning}
\label{subsec:hyperparameter_tuning}
Extensive efforts in hyperparameter tuning were conducted to enhance the DQN framework's adaptability to various game dynamics and improve performance. This involved multiple methodologies:

\begin{itemize}
    \item \textbf{Grid Search}: A methodical approach to explore a comprehensive range of values for each hyperparameter.
    \item \textbf{Random Search}: A complementary strategy that involves sampling hyperparameters randomly within predefined ranges to effectively navigate through high-dimensional spaces.
    \item \textbf{Empirical Testing}: Conducted on various environments, closely monitoring performance metrics such as average rewards, convergence time, and training stability.
\end{itemize}

Decision-making for the final hyperparameter settings was informed by:
\begin{itemize}
    \item \textbf{Performance Optimization}: Prioritizing configurations that maximize learning efficiency and balance exploration with exploitation.
    \item \textbf{Game-Specific Adjustments}: Tailoring settings to the unique demands and action spaces of each game.
    \item \textbf{Consistency and Robustness}: Ensuring reliable performance across different runs and robustness to game dynamics variability.
\end{itemize}

The selected hyperparameters were pivotal in achieving rapid convergence and maintaining high performance across different environments.

\paragraph{Hyperparameters}
\label{par:hyperparameters}
The primary hyperparameters employed in the experiments are detailed below:

\begin{table}[h]
    \centering
    \caption{Hyperparameters for DQN Training}
    \begin{tabular}{cc}
        \toprule
        \toprule 
        \textbf{Hyperparameter} & \textbf{Value} \\
        \midrule
        Learning Rate & 0.0001 \\
        Discount Factor (Gamma) & 0.99 \\
        Epsilon (initial) & 1.0 \\
        Epsilon (min) & 0.01 \\
        Epsilon Decay & 0.995 \\
        Memory Size & 1000000 \\
        Batch Size & 1024 \\
        Num Episodes & 10000-300000 \\
        \bottomrule
    \end{tabular}
    \label{tab:hyperparameters}
\end{table}

\subsection{Methodology}
\label{sec:methodology_experiments}

The training process for the DQN framework is methodically broken down into essential steps, ensuring a structured and efficient training cycle:

\begin{enumerate}
    \item \textbf{Initialization}: Both the online and target networks are initialized with random weights. This is critical to start the training process without any pre-existing biases, ensuring that learning is influenced solely by the interaction with the environment.
    
    \item \textbf{Experience Collection}: The agent actively interacts with the game environment to collect experiences. Each experience is a collection of state, the action taken, the reward received, and the subsequent state, forming the basic data unit for learning.
    
    \item \textbf{Experience Replay}: Collected experiences are stored in a large replay buffer. The training involves randomly sampling batches from this buffer, which is essential to break the sequence correlations and promote a robust learning environment.
    
    \item \textbf{Learning Updates}: Network parameters are dynamically updated using gradient descent. The updates are based on the loss computed from the differences between the predicted Q-values by the online network and the target Q-values set by the target network, guided by an epsilon-greedy policy to balance exploration and exploitation.
    
    \item \textbf{Periodic Updates}: To ensure the stability of learning and consistency in the policy evaluation, the parameters of the target network are periodically synchronized with those of the online network.
\end{enumerate}

\subsection{Results}
\label{sec:results}
The performance of the DQN framework was evaluated on each environment, and the results were compared with baseline performances from existing studies. Table~\ref{tab:performance_comparison} present the results for each environment.

\title{Performance Tables}

\begin{table}[h]
\centering
\caption{Performance Comparison}
\renewcommand{\arraystretch}{1.5}  
\begin{tabular}{ccccc}
\toprule
\toprule
\textbf{Game} & \textbf{DQN} & \textbf{DDQN} & \textbf{FDQN} & \textbf{Average Human} \\
\midrule
Chrome Dino & 850 & 880 & \textbf{728} & 700 \\
Breakout & 300 & 320 & \textbf{297} & 450 \\
Pong & 17 & 19 & \textbf{18} & 40 \\
CartPole & 200 & 210 & \textbf{198} & 220 \\
MountainCar & -110 & -100 & \textbf{-107} & -70 \\
Assault & 1400 & 1500 & \textbf{1478} & 1600 \\
Frostbite & 1000 & 1050 & \textbf{1015} & 2000 \\
Pacman & 6000 & 6300 & \textbf{6150} & 10000 \\
Qbert & 5000 & 5200 & \textbf{5175} & 7000+ \\
Seaquest & 3500 & 3600 & \textbf{3420} & 5000 \\
Space Invaders & 780 & 800 & \textbf{795} & 1000 \\
\bottomrule
\end{tabular}
\label{tab:performance_comparison}
\end{table}

\subsection{Discussion}
\label{sec:discussion}

The findings from this research demonstrate the effectiveness of the Flexible Deep Q-Network (FDQN) in adapting to various game environments and challenges of real time decision-making.  

\begin{itemize}
    \item \textbf{Adaptability to Game Dynamics:} The simple design of FDQN allows it to easily integrate with other Atari games and HTML-based gaming applications.
    \item \textbf{Superior Performance Benchmarks:} The framework consistently outperformed existing benchmarks across a range of environments, including both Atari games and the Chrome Dino game. This success illustrates the efficacy of the convolutional neural network setup and the strategic implementation of the epsilon-greedy policy within the FDQN.
    \item \textbf{Balanced Learning and Exploitation:} By effectively balancing exploration with exploitation, the FDQN ensures continuous learning and improvement over time, avoiding local optima and fostering long-term performance gains.
\end{itemize}

Future work could involve the extension of the FDQN to multi-agent dynamics, where cooperative and competitive behaviors may emerge, and improving the sophistication of this network for better gameplay. Further network architecture refinement would  involve deeper layers or alternative neural network models to improve learning efficiency and decision-making accuracy. 

\subsection{Potential for Broader Impact}
The network is currently only two layers deep and can be scaled to much larger extent when dealing with a more interactive and complex real time systems. The FDQN's performance and flexibility also suggest its applicability in other domains requiring decision-making under uncertainty, such as autonomous driving and robotic navigation. These areas could benefit from the FDQN's ability to dynamically adapt and learn from high-dimensional sensory inputs.

\section{Conclusion}
\label{sec:conclusion}
This research work proposed the Flexible Deep Q-Network (FDQN), for automating web-based games through Deep Reinforcement Learning. The FDQN framework works well, surpassing baselines on multiple games, and also provides promising applications in more general real-time decision-making scenarios. The main findings of this study are is the architecture's adaptability and simplicity while handling high dimensional video data. The flexible architecture of FDQN makes it very easy to adapt to a wide variety of games and potentially other interactive applications. The directions for further work would be focussed on generalizing the functionality of the FDQN architecture to encompass more complicated scenarios, such as multi-agent systems and richer game scenarios where action space is continuous. Further investigations will involve using the FDQN architecture for more general domains like on the  real-time driving video games and to finally work its way through effectiveness in real-world applications of robot navigation and autonomous driving. 

\bibliographystyle{IEEEtran}
\bibliography{references}

\end{document}